\documentclass{article}
\usepackage[preprint]{neurips_2026}
\usepackage[utf8]{inputenc}
\usepackage[T1]{fontenc}
\usepackage{hyperref}
\usepackage{url}
\usepackage{booktabs}
\usepackage{graphicx}
\usepackage{amsmath}
\usepackage{microtype}
\usepackage{xcolor}
\usepackage{float}

\title{Quantization Amplifies Determinism, Not Bias:\\Scale-Dependent Behavioral Effects of Serving-Time Weight Compression}

\author{%
  Dachi Kurtskhalia\\
  \texttt{dkurtskhalia@impel.ai}\\
}

\begin{document}

\maketitle

\begin{abstract}
Weight quantization largely determines the economics of serving open-weight
LLMs. Its costs are usually assessed with capability benchmarks, on which 4-bit
quantization of mid-sized models is often considered ``nearly free.'' We examine
a different question: when several answers are valid, does quantization change
what a model chooses to say? We serve three checkpoints (Qwen3-8B/14B/32B) at
three weight precisions (W4A16 AWQ, W8A16 FP8-Marlin, and bf16), holding the
hardware, software, and sampling configuration constant, and collect
approximately 71{,}000 completions paired by prompt and seed across two custom,
leak-checked prompt batteries. We pre-specified the analyses in three waves in
version control. At 8B, int4 reduces output diversity: the probability that two
samples for the same scenario recommend the same brand increases by 5.1
percentage points (prompt-paired sign-flip test, Holm $p{=}.023$; reproduced at
$+4.4$pp on a full regeneration of the arm), and lexical diversity falls
substantially (TTR $-0.011$, standardized effect $-0.51$; robust to a
length-controlled measure). At 14B and 32B, no content-concentration measure
reaches significance; instead, stylistic drift emerges (em-dash rate $+0.46$/1k
words at 14B and $+0.61$/1k at 32B, both Holm $p{\leq}.0024$). Pre-specified
tests of stereotype direction are null at every scale: outputs concentrate on
the modal answer for each prompt rather than on stereotypical answers.
Mechanistically, the token-level distribution becomes flatter (decision-token
entropy $+0.091$ bits, $p{=}.015$) while the semantic distribution, measured
directly from first-token log probabilities, becomes more concentrated
(collision $+2.6$pp, $p{=}.023$): individual tokens become less predictable
even as meanings become more repetitive. At 8B, the smallest size tested,
AWQ-int4 serving measurably narrows the range of suggestions; audits should
assess concentration as well as bias.
\end{abstract}

\section{Introduction}

Post-training weight quantization is one of the main ways open-weight models
become economical to deploy. For models in the 7--70B range, extensive evidence
from perplexity and capability benchmarks supports the prevailing view that W4
quantization has little measurable cost. Yet these evaluations largely concern
tasks with something close to a single correct answer. Deployed assistants more
often recommend, suggest, draft, or select defaults in settings where
\emph{many} answers are valid. In these settings, quality depends not only on
whether a model can produce an acceptable response, but also on \textbf{how
much of the valid answer space it continues to use}. An argmax-style evaluation
is not designed to detect damage at this distributional level.

Three lines of motivation converge here. RLHF and instruction tuning can
concentrate output distributions relative to those of base models; serving-time
compression may reinforce this tendency, leaving deployed models more
homogeneous than conventional benchmarks suggest. \citet{croq2026} show
that LLMs disproportionately volunteer Japan when asked to name ``a country'';
no previous work has tested whether compression intensifies this kind of
default-answer concentration. And in our motivating deployment (automotive
retail assistants served quantized for cost), two commercially important
failure modes are easy to conflate: \emph{bias} systematically favors
particular brands, while \emph{concentration} narrows the set of brands
recommended, limiting exposure of tail inventory. The two problems require
different audits and remedies. In our experiments, only the latter occurs.

\paragraph{Contributions.}
(1) A controlled serving-time precision ladder using the same official
checkpoint family at three precisions, with identical hardware, engine, flags,
and seeds, removing the provider confound common to API-based comparisons.
(2) A 3-size $\times$ 3-precision grid over two purpose-built batteries that
were adversarially reviewed and programmatically checked for leakage.
(3) A declare-then-run protocol: wave 1 pre-specified before any generation;
waves 2 and 3 after the 14B results were seen but before the 8B and 32B arms
were generated, so those arms carry clean declare-then-run status; the
repository history makes the sequence auditable.
(4) The findings: int4 reduces recommendation and lexical diversity at 8B;
these content effects disappear by 14B and are replaced by stylistic drift; no
stereotype amplification occurs at any scale; and token entropy rises even as
directly measured semantic entropy falls, indicating a shift from semantic
alternatives to surface variation rather than simple distribution sharpening.

\section{Related work}

GPTQ~\citep{frantar2022gptq}, AWQ~\citep{lin2024awq}, and
FP8~\citep{micikevicius2022fp8} are now standard serving methods. Across
conventional benchmarks, the accuracy literature generally finds W4
quantization to be nearly costless for mid-sized models, with most degradation
concentrated among smaller models~\citep{dettmers2023case};
quantization-induced increases in perplexity are well documented and align with
our token-level entropy results. Preference tuning is known to reduce output
diversity~\citep{kirk2024understanding,padmakumar2024does}, while lexicon
studies, including Kobak et al.'s work on ``excess
vocabulary''~\citep{kobak2024delving}, identify a recognizable LLM style
whose markers we adopt directly. \citet{croq2026} document a tendency to
default to Japan in prompts that ask models to name a country; we adopt their
elicitation paradigm and study the interaction with compression for the first
time. Finally, the recommender-systems literature treats aggregate diversity
and catalog coverage as primary evaluation
criteria~\citep{ziegler2005improving,kunaver2017diversity}; our
collision-probability measure is an LLM-oriented analogue of intra-list or
aggregate diversity.

\section{Method}

\paragraph{Precision ladder.}
We evaluate three official Qwen3 checkpoints~\citep{qwen3-2025} (8B, 14B, and
32B), serving each at three weight precisions from official releases:
\textbf{W4A16} (AWQ), \textbf{W8A16} (the official FP8 checkpoint, executed
weight-only through Marlin kernels~\citep{frantar2024marlin} on Ampere GPUs,
with activations remaining in bf16), and \textbf{bf16}. This design produces a
monotone \emph{weight-precision} series while holding activation precision
constant, though the rungs differ in quantization method as well as bit-width.
We serve one model at a time with vLLM 0.11.0~\citep{kwon2023vllm}, using
identical flags across all precision tiers within a model; tensor parallelism
remains constant within each model (TP=4 for 8B/14B, TP=8 for 32B). We sample
at temperature 0.8 and top-p 0.95 with a maximum of 400 tokens, and for each
prompt and precision tier we draw \textbf{n=20 samples using the same set of
seeds across tiers}, removing sampling-RNG variation from every contrast;
residual nondeterminism from dynamic batching remains, is symmetric across
tiers, and is exercised deliberately by the regeneration check. Total compute
cost was approximately \$80.

\paragraph{Batteries.}
The cultural battery contains 218 country-eliciting prompts spanning 11
domains, all of which ask the model to supply a country without mentioning one
in the prompt, plus 80 style-control prompts; each cultural prompt carries a
construction-time \texttt{country\_pull} label (none: 110, weak: 65, strong:
43, with prompt-specific targets; 30 of the 43 strong pulls target Japan). The
recommendation battery contains 96 brand-blind car-buying scenarios across
eight buyer domains; every prompt requires the answer to identify a brand but
mentions no brand itself, with \texttt{brand\_pull} labels (none: 72, weak: 16,
strong: 8). Both batteries were produced by parallel generator lanes, reviewed
by adversarial verifier lanes, and independently checked for leakage against
the gazetteers: no prompt text contains a country, city, nationality, brand, or
model name.

\paragraph{Measurement.}
Countries are extracted with a 78-country gazetteer and context-aware
disambiguation rules; brands with a 57-make gazetteer whose four ambiguity
classes distinguish vehicle brands from phrases such as ``Harrison Ford,'' ``my
aunt Mercedes,'' ``4GB of RAM,'' and ``the book of Genesis'' (30-case
adversarial test suite). Style metrics are the Kobak focal-word rate, em-dash
rate, ``not X but Y'' constructions, and type-token ratio. No rows were dropped
from the final grid, and more than 99\% of completions ended with a natural
stop.

\paragraph{Statistical protocol.}
The \textbf{prompt} is our unit of analysis. All primary tests compare paired
int4$-$bf16 contrasts at the prompt level, using two-sided sign-flip
permutation tests (10{,}000 flips) and prompt-cluster bootstrap confidence
intervals (5{,}000 resamples). Corpus-level quantities are computed from
mentions rarefied to the minimum count shared across tiers, with Miller--Madow
estimates~\citep{miller1955note} as a robustness check; we treat CI-only
quantities as supporting evidence throughout, with confirmatory claims resting
solely on the Holm-corrected permutation tests. Wave 1 covers scalar style and
content metrics; wave 2 measures within-prompt concentration through the
collision probability of the first-mentioned entity and mass on the top five
bf16 entities; wave 3 tests stereotype direction through stereotype-set share
and pull-target compliance, with scenario sets defined from prior knowledge,
not from observed tier differences. Holm--Bonferroni
correction~\citep{holm1979simple} applies within each declared family; no tests
were added after inspecting the corresponding data.

\paragraph{Mechanism instruments.}
A separate 8B run records top-20 log-probability summaries per generated token.
In addition, because each of the 57 makes begins with a unique first token
under the Qwen3 tokenizer, we measure each prompt's brand distribution
directly: we append the fixed assistant prefix ``The brand I would recommend
is,'' request one token with top-50 log probabilities, and map brand-initial
tokens to makes; the measurement is exact conditional on the fixed decision
frame and top-50 truncation, with brand-initial tokens carrying 24.0\% (bf16) and 26.2\% (int4)
of next-token mass and metrics computed on the renormalized brand mass.

\section{Results}

\subsection{At 8B, int4 collapses recommendation diversity}

\begin{table}[t]
\centering
\caption{8B, int4 $-$ bf16. CIs are prompt-cluster bootstrap 95\%.}
\label{tab:8b}
\small
\begin{tabular}{lccc}
\toprule
measure & $\Delta$ & 95\% CI & $p$ (Holm) \\
\midrule
brand collision probability (wave 2) & $\mathbf{+0.051}$ & $[+0.012, +0.090]$ & .012 (.023) \\
bf16-top-5 brand mass (wave 2) & $+0.014$ & $[-0.008, +0.038]$ & CI spans 0 \\
rarefied brand entropy (wave 1, supporting) & $-0.021$ & $[-0.041, -0.001]$ & CI excludes 0 \\
type-token ratio (wave 1) & $\mathbf{-0.0113}$ & $[-0.0141, -0.0083]$ & $<10^{-4}$ ($<4{\times}10^{-4}$) \\
MATTR-100 (length-robust check) & $\mathbf{-0.0066}$ & $[-0.0090, -0.0042]$ & $<10^{-4}$ \\
\bottomrule
\end{tabular}
\end{table}

\begin{figure}[t]
\centering
\includegraphics[width=\linewidth]{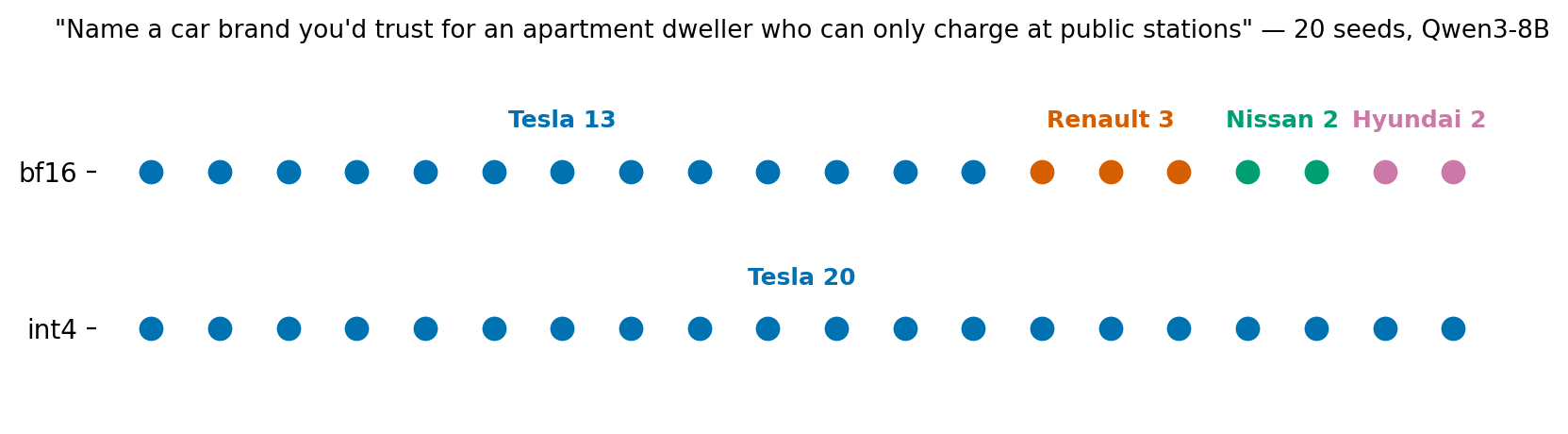}
\caption{The largest observed per-prompt shift (Qwen3-8B, same 20 seeds per
tier): bf16 recommends four distinct brands; int4 recommends Tesla in all 20
samples.}
\label{fig:exhibit}
\end{figure}

Because TTR mechanically decreases with length and int4 completions are longer
at 8B ($+5.7$ words, $p{<}10^{-4}$), we verify the effect on
MATTR-100~\citep{covington2010mattr}, a moving-window measure independent of
length: it survives essentially unchanged. We reproduce the collision effect on
a full regeneration of the arm ($+4.4$pp, $p{=}.029$). The effect is
battery-specific and heterogeneous: the cultural battery shows no 8B
concentration on any measure (country collision $-1.2$pp, n.s.), and the median
per-prompt collision delta is 0 (IQR $[0, +0.14]$): most scenarios do not
move; a substantial minority collapse hard. The largest observed per-prompt
shift, using the same 20 seeds (Figure~\ref{fig:exhibit}): \emph{``Name a car
brand you'd trust for an apartment dweller who can only charge at public
stations.''} With bf16, the model recommends Tesla 13 times, Renault 3 times,
Nissan 2 times, and Hyundai 2 times. With int4, it recommends \textbf{Tesla in
all 20 samples}. The collapse is \textbf{modal rather than global}: Toyota's
overall first-mention share rises from 42\% to 47\% while Honda's falls from
14\% to 9\%, and their combined share remains unchanged: the modal answer
for a given scenario absorbs probability from that scenario's alternatives, but
the identity of the mode varies across scenarios.

\subsection{The effects dissociate by scale}

\begin{table}[t]
\centering
\caption{int4 $-$ bf16 across model sizes. $\dagger$: declared outside the
wave-1 Holm family; raw $p$ shown elsewhere in the repository.}
\label{tab:scale}
\small
\begin{tabular}{lccc}
\toprule
int4 $-$ bf16 & 8B & 14B & 32B \\
\midrule
brand collision & $\mathbf{+.051}$ ($p{=}.012$) & $-.007$ (n.s.) & $-.023$ (n.s.) \\
rarefied brand entropy & $-.021$ (CI$<$0) & $-.002$ (n.s.) & $+.004$ (n.s.) \\
TTR (cultural) & $\mathbf{-.0113}$ ($p{<}10^{-4}$) & $-.0026$ ($p{=}.022$, Holm .058) & $+.0003$ (n.s.) \\
em-dash /1k (cultural) & $+.26$ (n.s.) & $\mathbf{+.46}$ (Holm .002) & $\mathbf{+.61}$ (Holm .0024) \\
\bottomrule
\end{tabular}
\end{table}

\begin{figure}[t]
\centering
\includegraphics[width=\linewidth]{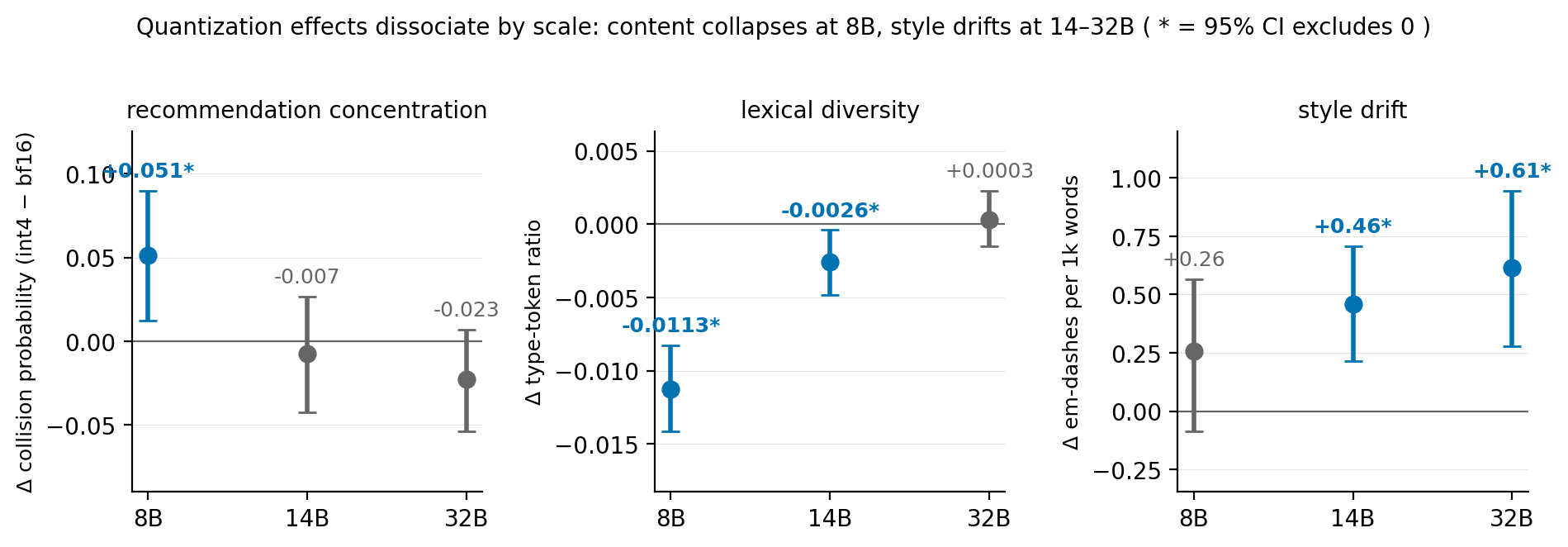}
\caption{Paired int4 $-$ bf16 deltas with 95\% prompt-cluster bootstrap CIs by
model size. Content concentration appears only at 8B; stylistic drift only at
14B and 32B.}
\label{fig:dissociation}
\end{figure}

The content effects are confined to the smallest model
(Figure~\ref{fig:dissociation}). By contrast, the em-dash drift is significant
only at 14B and 32B: $+0.46$ and $+0.61$ per 1k words in absolute terms
(near-identical standardized effects of 0.25), corresponding to $+19\%$ and
$+12\%$ relative to each model's own baseline. The drift is significant for the
cultural prompts but not for the placeless style-control arm at any model size
(Appendix~D). Thus, the measured effect of quantization does not simply
diminish as models grow. Its form changes: the 8B model becomes less diverse in
\emph{what} it says, whereas the larger models retain their content
distribution but shift in \emph{how} they express it.

\subsection{No evidence of stereotype amplification (pre-specified)}

All three wave-3 measures are null at all three sizes (stereotype-set share
$-.011$ to $-.017$; brand pull-target compliance $-.007$ to $+.017$; $n{=}23$--$24$
for the pull-compliance test, so its nulls bound only large effects; full
tables in Appendix~C). The only significant wave-3 result points in the
opposite direction: at int4, the 32B model complies \textbf{less} with weak
country stereotypes ($-4.4$pp, Holm $p{=}.043$). Combined with \S4.1, these
findings show concentration without a consistent stereotypical direction. Even
within the same stereotype set, Toyota gains share while Honda loses it.
\textbf{Quantization amplifies determinism, not bias.} For countries, we
observe small and mixed movements only at 32B (unprompted Japan share
$+2.3$pp, provoked Japan share $-3.2$pp); because these directions do not
support a coherent amplification account, we treat them as suggestive evidence
of redistribution and do not interpret them further.

\subsection{Mechanism: token noise increases while semantic diversity declines}

\begin{figure}[t]
\centering
\includegraphics[width=\linewidth]{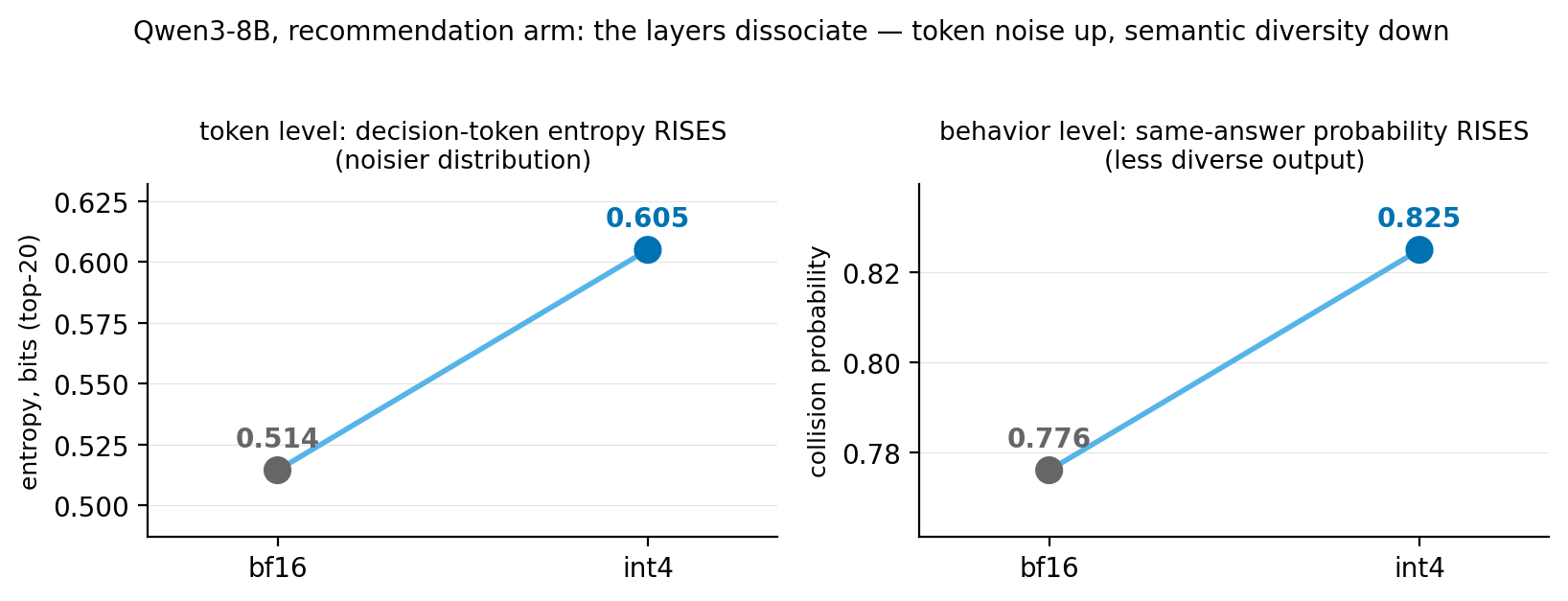}
\caption{Qwen3-8B, recommendation arm: token-level entropy rises under int4
while behavioral same-answer probability also rises; the two description
levels dissociate.}
\label{fig:mechanism}
\end{figure}

Under int4 at 8B, the token-level distribution becomes \emph{flatter}:
decision-token entropy increases by \textbf{0.091 bits} (CI $[+0.020, +0.163]$,
$p{=}.015$), top-1 probability falls by 3.8pp ($p{=}.024$), and the share of
near-forced tokens declines ($p{=}.009$). Despite this, the brand distribution
itself becomes more \emph{concentrated}: exact collision increases by
\textbf{2.6pp} (CI $[+0.4, +4.7]$, $p{=}.023$; excluding short-prefix makes,
$p{=}.019$), with a mean per-prompt Jensen--Shannon divergence of 0.040 bits
between the tiers (Figure~\ref{fig:mechanism}). Total brand mass itself rises
slightly at int4 ($24.0\%{\to}26.2\%$), so the concentration is not an artifact
of shrinking brand coverage. We therefore measure both sides of the
dissociation directly: the additional token-level uncertainty appears as
variation in surface form while semantically distinct alternatives lose
probability mass. One practical consequence follows: \textbf{this failure mode
cannot be audited from token-level logits alone.}

\subsection{Robustness}

At temperature 0.4, the TTR decline persists ($-0.0090$, $p{<}10^{-4}$); the
collision contrast retains the same direction but falls below significance
($+3.2$pp, $p{=}.14$) as colder sampling concentrates both precision tiers
(paired mean collision 0.901/0.933 versus 0.776/0.827 at temperature 0.8),
leaving roughly half the headroom for any between-tier separation. We interpret
the attenuation as ceiling compression rather than evidence that the effect
vanishes.

\section{Discussion}

A generation that begins with a non-modal token must sustain a coherent
trajectory to produce a non-modal answer. Quantization noise may degrade the
model's representation of alternative answers more severely than its
representation of the mode; under this account, noisy initial tokens are more
likely to converge on the modal answer later in the sequence. Our results
establish the dissociation itself; this explanation remains interpretive.

Recommendation audits typically measure \emph{skew}: which options the model
favors. A small quantized model can pass every stereotype-direction audit we
ran while still reducing the set of answers it provides for a given query;
ours did. Deployment audits should therefore include concentration measures
such as collision probability or the number of distinct answers across $k$
repeated queries. At 8B, the smallest size tested, AWQ-int4 serving
measurably narrows the range of recommendations; the effect is absent at 14B
and 32B in this family. Where small quantized models must support
recommendations, practitioners should consider a larger model at the same
precision, diversity-promoting decoding, or conditioning explicitly on
available inventory.

\section{Limitations}

We examine only one model family, Qwen3, so the apparent boundary near 14B may
be family-specific, and only one quantizer at each precision; calibration-set
effects remain untested and are the subject of a pre-specified follow-up. Our
W8 condition runs weight-only on Ampere and does not represent W8A8 serving.
The recommendation battery covers a single domain (automotive retail with
US-market framing) and the main analysis is limited to English. 32B was
served at TP=8 on a different instance type, so cross-scale comparisons
involving 32B carry a parallelism difference alongside the scale difference.
The 8B concentration effect is heterogeneous across prompts (median per-prompt
delta 0), so it should be read as ``a minority of scenarios collapse hard,''
not as a uniform shift. Finally, the collision contrast falls below
significance at temperature 0.4.

\section*{Reproducibility}

All code, prompts with construction provenance, gazetteers with adversarial
test suites, per-wave analysis scripts, frozen statistics CSVs, and figures are
maintained in a version-controlled repository whose history records the
declare-then-run sequence; it will be released publicly with the camera-ready
version. Scoring and analysis run locally against any OpenAI-compatible vLLM
endpoint.

\section*{Acknowledgments and disclosure of AI assistance}

Compute for the experiments was provided by Impel. We thank Ana Kolkhidashvili
for reviewing the manuscript. All experiment launches, budget decisions, and
scientific judgment calls were made by the author. Prompt batteries, analysis code, and manuscript drafts were developed
with substantial assistance from AI systems (Anthropic Claude; prose editing
with OpenAI Codex); all analyses were pre-specified in version control with
auditable ordering, all reported statistics derive from frozen data files in
the repository, and the author has verified and takes responsibility for every
claim.

\bibliographystyle{plainnat}
\bibliography{refs}

\appendix

\section{Complete wave-1 results}

The full 30-row wave-1 table (Kobak, em-dash, not-X-but-Y, TTR,
mentions/completion, rarefied entropies with Miller--Madow values, Japan
shares, and Toyota+Honda share for all three models) is provided in the
accompanying repository (\texttt{results/stats\_summary.csv}); headline rows
appear in Tables~\ref{tab:8b} and~\ref{tab:scale}.

\section{Complete wave-2 and exploratory concentration results}

\begin{table}[H]
\centering
\caption{Wave-2 confirmatory and exploratory concentration measures, int4 $-$ bf16, all sizes.}
\small
\begin{tabular}{llcccc}
\toprule
model & metric & $n$ & $\Delta$ & 95\% CI & $p$ (Holm) \\
\midrule
8B & collision, first brand & 91 & $+0.051$ & $[+0.012, +0.090]$ & .012 (.023) \\
8B & bf16-top5 mass, brands & --- & $+0.014$ & $[-0.008, +0.038]$ & --- \\
8B & collision, first country & 183 & $-0.012$ & $[-0.043, +0.020]$ & .452 (.452) \\
8B & bf16-top5 mass, countries & --- & $-0.003$ & $[-0.032, +0.027]$ & --- \\
8B & support size, brand (of 20) & 91 & $-0.176$ & $[-0.352, 0.000]$ & .068 (expl.) \\
8B & support size, country (of 20) & 190 & $+0.016$ & $[-0.189, +0.221]$ & .925 (expl.) \\
8B & strong-pull compliance, countries & 35 & $-0.038$ & $[-0.095, +0.007]$ & .191 (expl.) \\
14B & collision, first brand & 89 & $-0.007$ & $[-0.042, +0.027]$ & .676 (.676) \\
14B & bf16-top5 mass, brands & --- & $+0.002$ & $[-0.021, +0.025]$ & --- \\
14B & collision, first country & 190 & $-0.020$ & $[-0.050, +0.010]$ & .197 (.393) \\
14B & bf16-top5 mass, countries & --- & $+0.001$ & $[-0.021, +0.022]$ & --- \\
14B & support size, brand (of 20) & 91 & $+0.066$ & $[-0.099, +0.231]$ & .531 (expl.) \\
14B & support size, country (of 20) & 196 & $+0.051$ & $[-0.117, +0.219]$ & .590 (expl.) \\
14B & strong-pull compliance, countries & 39 & $+0.011$ & $[-0.015, +0.050]$ & .811 (expl.) \\
32B & collision, first brand & 92 & $-0.023$ & $[-0.054, +0.007]$ & .170 (.340) \\
32B & bf16-top5 mass, brands & --- & $-0.005$ & $[-0.035, +0.022]$ & --- \\
32B & collision, first country & 196 & $-0.012$ & $[-0.036, +0.011]$ & .314 (.340) \\
32B & bf16-top5 mass, countries & --- & $+0.020$ & $[+0.003, +0.040]$ & --- \\
32B & support size, brand (of 20) & 92 & $+0.109$ & $[-0.043, +0.261]$ & .229 (expl.) \\
32B & support size, country (of 20) & 202 & $+0.005$ & $[-0.153, +0.173]$ & 1.000 (expl.) \\
32B & strong-pull compliance, countries & 41 & $-0.023$ & $[-0.055, +0.001]$ & .123 (expl.) \\
\bottomrule
\end{tabular}
\end{table}

\section{Complete wave-3 (stereotype-direction) results}

\begin{table}[H]
\centering
\caption{Wave-3 stereotype-direction measures, int4 $-$ bf16, all sizes.}
\small
\begin{tabular}{llcccc}
\toprule
model & metric & $n$ & $\Delta$ & 95\% CI & $p$ (Holm) \\
\midrule
8B & brand stereotype-set share & 91 & $-0.014$ & $[-0.034, +0.006]$ & .193 (.578) \\
8B & brand pull compliance & 23 & $+0.017$ & $[-0.000, +0.052]$ & 1.000 (1.000) \\
8B & country pull compliance (weak) & 55 & $-0.017$ & $[-0.054, +0.008]$ & .390 (.779) \\
14B & brand stereotype-set share & 91 & $-0.017$ & $[-0.047, +0.010]$ & .247 (.741) \\
14B & brand pull compliance & 23 & $-0.007$ & $[-0.055, +0.033]$ & 1.000 (1.000) \\
14B & country pull compliance (weak) & 58 & $+0.006$ & $[-0.023, +0.037]$ & .737 (1.000) \\
32B & brand stereotype-set share & 92 & $-0.011$ & $[-0.034, +0.012]$ & .372 (.745) \\
32B & brand pull compliance & 23 & $+0.001$ & $[-0.020, +0.020]$ & 1.000 (1.000) \\
32B & country pull compliance (weak) & 60 & $\mathbf{-0.044}$ & $[-0.083, -0.011]$ & .014 (.043) \\
\bottomrule
\end{tabular}
\end{table}

\section{Style-control arm (placebo check)}

The em-dash and Kobak drifts are tested on the 80 placeless style-control
prompts; none reaches significance at any size, supporting the specificity of
the cultural-prompt drift (no formal arm-by-tier interaction test was run).

\begin{table}[H]
\centering
\caption{Style-control (placebo) arm: Kobak and em-dash deltas on the 80 placeless prompts.}
\small
\begin{tabular}{llccc}
\toprule
model & metric & $\Delta$ & 95\% CI & $p$ \\
\midrule
8B & Kobak /1k & $-0.179$ & $[-0.506, +0.143]$ & .304 \\
8B & em-dash /1k & $+0.252$ & $[-0.480, +0.925]$ & .500 \\
14B & Kobak /1k & $+0.189$ & $[-0.214, +0.577]$ & .359 \\
14B & em-dash /1k & $-0.223$ & $[-0.771, +0.356]$ & .455 \\
32B & Kobak /1k & $+0.408$ & $[-0.036, +0.843]$ & .070 \\
32B & em-dash /1k & $+0.458$ & $[-0.149, +1.072]$ & .161 \\
\bottomrule
\end{tabular}
\end{table}

\end{document}